\documentclass[journal]{IEEEtran}

\usepackage{floatrow}
\usepackage{caption}
\usepackage[pdftex]{graphicx}
\DeclareGraphicsExtensions{.drawio.pdf,.pdf,.jpeg,.png}

\usepackage{amssymb}
\usepackage{gensymb}
\usepackage{amsmath}
\usepackage{subfigure}
\usepackage{cite}
\usepackage[flushleft]{threeparttable}
\usepackage{algorithm}
\usepackage{algpseudocode}

\DeclareMathOperator*{\argmin}{argmin}

\begin{document}

\title{Semi-Synthetic Dataset Augmentation for Application-Specific Gaze Estimation}

\author{C. Leblond-Menard, G. Picard-Krashevski and S. Achiche}

\maketitle

\begin{abstract}
    Although the number of gaze estimation datasets is growing, the application of appearance-based gaze estimation methods is mostly limited to estimating the point of gaze on a screen. This is in part because most datasets are generated in a similar fashion, where the gaze target is on a screen close to camera's origin. In other applications such as assistive robotics or marketing research, the 3D point of gaze might not be close to the camera's origin, meaning models trained on current datasets do not generalize well to these tasks. We therefore suggest generating a textured tridimensional mesh of the face and rendering the training images from a virtual camera at a specific position and orientation related to the application as a mean of augmenting the existing datasets. In our tests, this lead to an average 47\% decrease in gaze estimation angular error.

\end{abstract}

\section{Introduction}

In recent years the number of gaze estimation datasets is growing \cite{fischerRTGENERealTimeEye2018,krafkaEyeTrackingEveryone2016,suganoLearningbySynthesisAppearanceBased3D2014,zhangItWrittenAll2017,zhangMPIIGazeRealWorldDataset2019}. However, the application of appearance-based gaze estimation methods remains challenging due to the range of the gaze angles and head poses within those. While some datasets have a very large distribution of gaze angles and head poses, gaze estimation models trained on those still tend to be much less accurate when compared to datasets with a much limited distribution \cite{chengAppearancebasedGazeEstimation2021}.

To begin with, the gaze angles are the pitch and yaw angles of the gaze direction vector, which is the normalized vector between the gaze origin, generally defined as either an eye or the middle point between the eyes, and the point of gaze of a user. As for the head pose, it is generally defined as a 3D rotation matrix which describes the rotation of the head of the user with respect to the camera's reference frame.

As such, as the range of gaze angles and head poses of the dataset's samples increases, the problem complexity also increases and a learning-based model for gaze estimation will have to learn a more complex distribution, which generally leads to lower accuracy.

In certain applications, a large gaze angle and head pose range might not be relevant, such as is the case for assistive robotics. Indeed, in these cases, the gaze distribution might be limited to a specific forward facing region with respect to the user. For example, the allowed gaze direction for object selection in an assistive robotics task such as automated grasping might be limited to a box region in front of the user which corresponds to the robot reach \cite{cioProofConceptAssistive2019}.  Although the gaze distribution might have a similar range width or variance in this application as most available gaze estimation datasets, the center or mean of the region of gaze might be very different.

\begin{figure}
\subfigure[Gaze pitch and yaw]{\includegraphics[width=.4\linewidth]{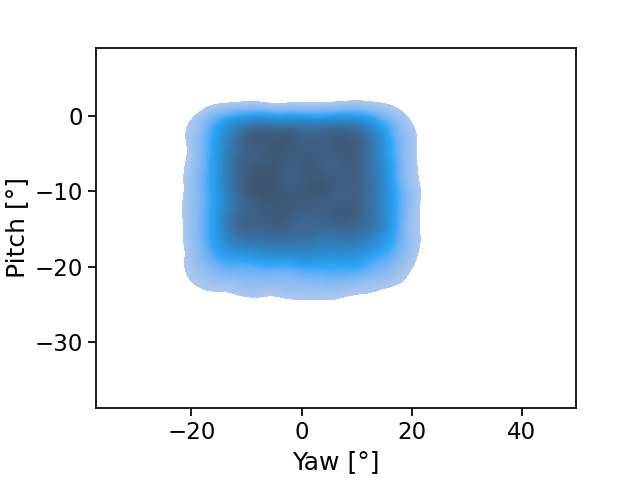}}
\hfill
\subfigure[Head pose pitch and yaw]{\includegraphics[width=.4\linewidth]{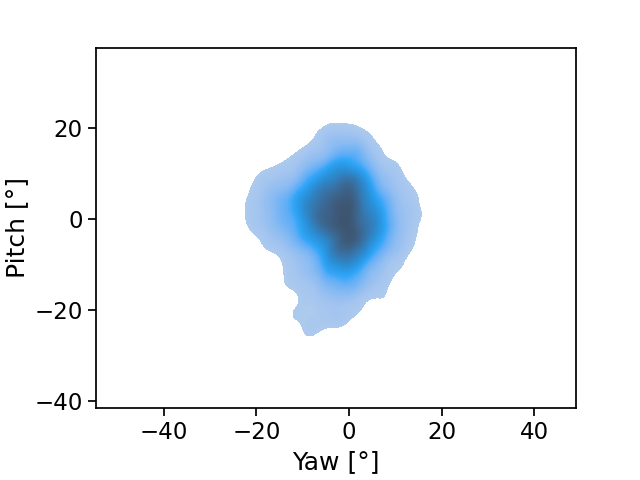}}
\caption{Distribution of the gaze direction and head pose pitch and yaw with respect to the camera's center axis for the MPIIFaceGaze dataset \cite{zhangItWrittenAll2017}.}
\floatfoot{Note that the gaze distribution is completely skewed toward looking under the camera. While this is representative of a scenario where we try to find the gaze point on a screen with a webcam above it, our sample application would lead to the complete opposite, where the user is looking almost exclusively above the camera.}
\label{fig:distribution_mpiifg}
\end{figure}

In fact, most datasets currently available were created from images taken from a webcam-type camera, generally fixed on a screen, laptop, tablet or phone, while the user is looking at a target shown on the screen or device \cite{krafkaEyeTrackingEveryone2016,zhangItWrittenAll2017,zhangMPIIGazeRealWorldDataset2019}. As can be seen from the distribution presented in Fig. \ref{fig:distribution_mpiifg}, this limits the distribution of gaze angles to regions close to or below the camera. While this is an accurate scenario for gaze estimation models where the goal is to estimate the point of gaze on a screen, it does not translate well to other applications where the user might not be looking directly toward the camera in 3D space.

On the other hand, other datasets \cite{suganoLearningbySynthesisAppearanceBased3D2014,zhangETHXGazeLargeScale2020,fischerRTGENERealTimeEye2018,kellnhoferGaze360PhysicallyUnconstrained2019} have tried to expand the limited distribution of the gaze angles in datasets generated from point on the screen by placing multiple cameras at several positions around the user \cite{suganoLearningbySynthesisAppearanceBased3D2014,zhangETHXGazeLargeScale2020} or by placing the user at random positions \cite{fischerRTGENERealTimeEye2018,kellnhoferGaze360PhysicallyUnconstrained2019}. While this might solve the issue of the limited distribution, it remains challenging to train on and can lead to poor accuracy due to the limited number of samples spread over a large range of gaze angles and head poses \cite{chengAppearancebasedGazeEstimation2021}.

To mitigate this issue, we suggest generating a mesh of the detected faces of the users in the dataset and then training a gaze estimation model by using pose transformed version of the face meshes and corresponding gaze directions. This is done through methods similar to what has already been done to augment the span of the gaze angles range \cite{qinLearningbyNovelViewSynthesisFullFaceAppearanceBased2022,suganoLearningbySynthesisAppearanceBased3D2014}, but these methods are adapted to instead improve the application-specific gaze estimation accuracy, e.g., assistive robotics tasks such as automated grasping. We also provide a simple way of generating the face mesh without having to use multiple viewpoints of the same sample through using multiple cameras as done in \cite{suganoLearningbySynthesisAppearanceBased3D2014}.

\begin{figure}
\centering{\includegraphics[width=\columnwidth]{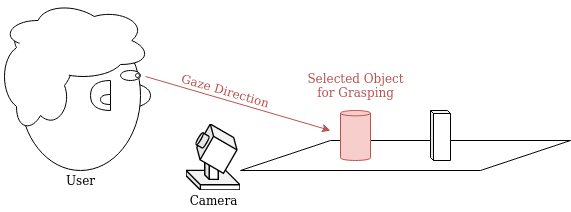}}
\caption{Sample application of gaze estimation for object selection for automatic grasping in assistive robotics.}
\floatfoot{As a sample application, we use the task of object selection to be grasped by an assistive robotic arm. In this scenario, a camera is used to estimate the gaze direction of the user to select an object. Once an object is selected, an assistive robotic arm (not shown) mounted close to the user, e.g. on their motorized wheelchair, moves in to grasp the object. In this application, the camera is mounted in front of the user typically on their motorized wheelchair, but behind the scene ahead and its objects. It is angled upward to look at the user without blocking the user's view.}

\label{fig:vis_scene}
\end{figure}

As later demonstrated in this paper, this allows to reach a much higher application-specific gaze estimation accuracy. We will use an assistive robotic arm controlled via gaze estimation as an example application in this paper. See \cite{cioProofConceptAssistive2019} for more details about the application and refer to Fig \ref{fig:vis_scene} for a visualization and explanation.

\section{Methodology}
The methodology used in this paper follows 8 steps:
\begin{enumerate}
  \item Align the face of each dataset sample to a face model.
  \item Solve for the pose of the face under a perspective projection using the obtained aligned facial landmarks.
  \item Generate the textured mesh of the face.
  \item Apply the inverse pose transform to the gaze direction to make it correspond with the generated centered face mesh.
  \item Compute the sampling distribution given the expected pose of the head and its variance for the application.
  \item Apply pose transforms of the head to the samples by sampling from the head pose distribution.
  \item Project the face mesh to virtual normalized camera plane for each eye.
  \item Use the transformed samples as input for training a gaze estimation model.
\end{enumerate}

\begin{figure}
  \centering{\includegraphics[width=\columnwidth]{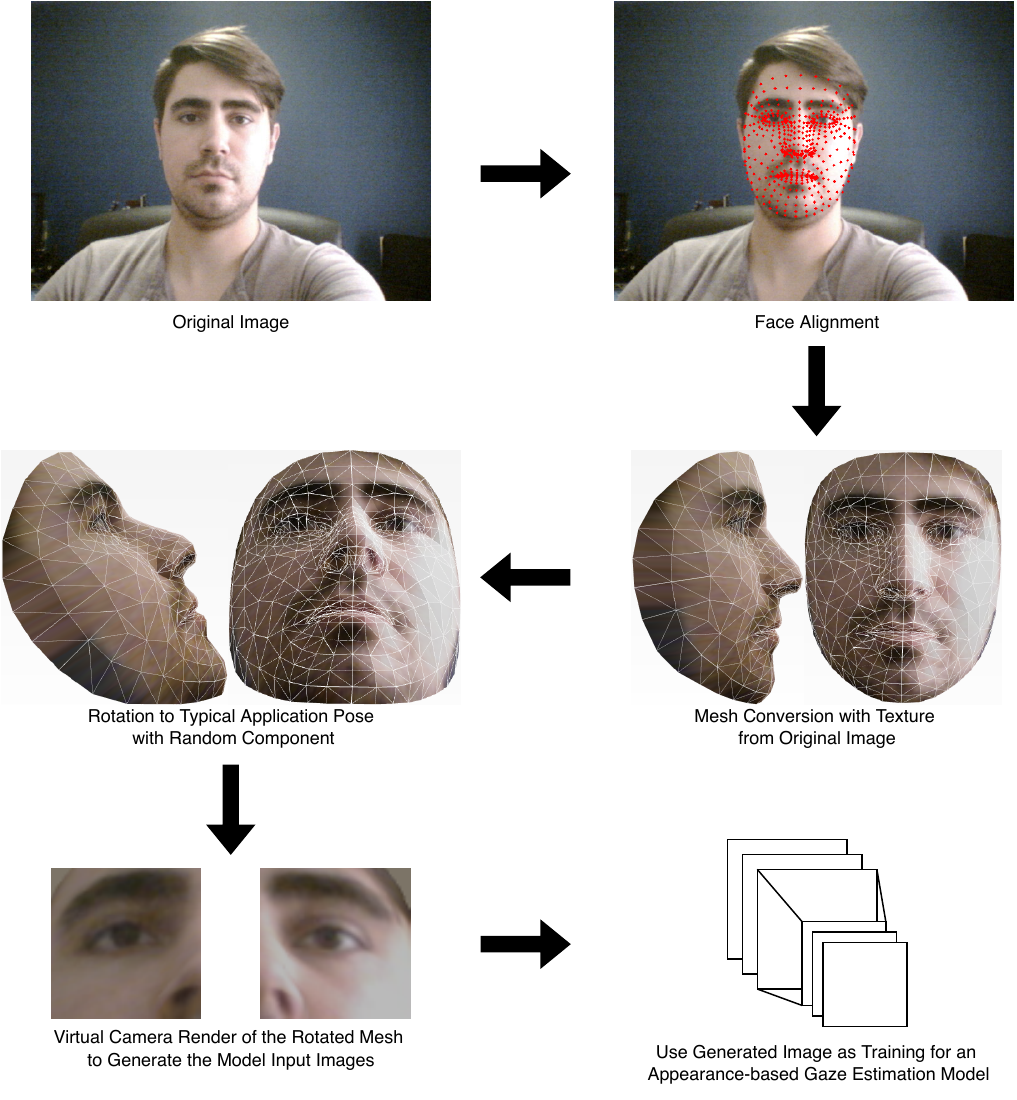}}
  \caption{Simplified visual workflow of the augmentation method with real example images.}
  \floatfoot{The red points and wireframe on the meshes were added for visualization purposes and are not part of the actual workflow. It should also be noted that the rotation demonstrated has no random component, but instead is a rotation of exactly 30 degrees in pitch and 0 degrees in yaw, which is the mean of the random rotation used in this paper, as presented in Eq. \ref{eq:bivariate_normal_aug}.
    The person visible in the image has given their consent to appear in this paper.}

  \label{fig:augment_workflow}
\end{figure}

Refer to Figure \ref{fig:augment_workflow} for a visualization of the workflow with respect to the modification made to the original image. The following sections will delve into the details of the previously described steps.

\subsection{Facial Alignment} \label{sect:facial_align}

Using any facial alignment method \cite{alvarezcasadoRealtimeFaceAlignment2021} for which the base face model includes 3D coordinates, align each face of the dataset from the corresponding sample image. In our case, we used the model developed by \cite{kartynnikRealtimeFacialSurface2019}, but any accurate model can do.

After running the face alignment model, a set of 2D points in pixels is obtained. These points correspond to facial features or areas as learned by the facial alignment model. We will call the set of obtained 2D points in screen (pixels) coordinates $\mathbf{X_s}$.

\subsection{Pose Under Perspective Projection}

Because the previously obtained face points $\mathbf{X_s}$ are in 2D screen coordinates, they cannot be directly used to find the pose of the head between the base pose given by the face model and the actual pose in the computed sample. Indeed, algorithms developed to solve the orthogonal Procrustes problem cannot be used as is to solve the problem under projective transformation. For such a case, several methods have been proposed and the most common ones have been implemented under OpenCV's \verb"solvePnP" function \cite{OpenCVPerspectivenPointPnP}. This function finds the best $\mathrm{SE}(3)$ transformation (simultaneous rotation and translation in 3D) which, when projected according to the sample's camera matrix and distortion coefficient, minimizes the error between the input face 2D points and the computed face model 3D points under the pose and projection transforms. Simply put, we find the transformation $\mathbf{^bT_a}$ which maps the base face model 3D points to a new set of 3D points corresponding to the actual pose of the face in the sample:

\begin{equation}
  \mathbf{^bT_a} = \underset{\mathbf{T} \in \mathrm{SE}(3)}{\argmin} || \mathbf{\hat{X}_s} - \mathbf{X_s} ||^2 = \underset{\mathbf{T} \in \mathrm{SE}(3)}{\argmin} || \mathbf{C \  T \  X_m} - \mathbf{X_s} ||^2
\end{equation}

$\mathbf{\hat{X}_s}$ corresponds to the face points obtained by applying the estimated transformation $\mathbf{T}$ to the base face model points $\mathbf{X_m}$ and projecting those using the camera matrix $\mathbf{C}$. It is assumed that the screen-space coordinates $\mathbf{X_s}$ and $\mathbf{\hat{X}_s}$ and the face model 3D coordinates $\mathbf{X_m}$ are expressed in homogenous coordinates, namely $[u, v, 1]$ and $[X, Y, Z, 1]$, and as such have been normalized as to have their last component normalized to 1.

\subsection{Textured Face Mesh Generation}
To generate the textured face mesh, we need:
\begin{itemize}
  \item A set of 3D vertices.
  \item A set of mesh faces defined by 3 or more of the previously defined vertices.
  \item A texture image.
  \item A $(u,v)$ coordinate on the texture image for each 3D vertex.
\end{itemize}

\textit{It is worth noting that the mesh faces is the name given to the polygonal faces of a mesh, not a human face. These mesh faces are generally triangles.}

The set of 3D vertices can be defined as the model face points $\mathbf{X_m}$. The texture image can simply be the input face image used in the alignment step (step 1). When creating the mesh's texture using the face image, the $(u,v)$ texture coordinates to associate to each vertex are $\mathbf{X_s}$, the facial landmark positions in screen space as obtained during the alignment.

The main difficulty here is to generate the faces from the set of 3D vertices if the base face model does not contain any mesh face definitions, which is not the case for the method  we used \cite{kartynnikRealtimeFacialSurface2019}. If the mesh faces definitions are not included, we suggest using surface reconstruction methods such as the Ball-Pivoting Algorithm \cite{bernardiniBallPivotingAlgorithmSurface1999a} to generate the set of mesh faces associated with the vertices, which are themselves the facial landmarks.

\subsection{Inverse Pose Transform Correction to Gaze Direction}

As the gaze annotations from the datasets are given with respect to the actual pose of the face and not the centered, base face model pose, we need to correct the gaze direction vector using the previously computed pose. Given the actual gaze direction as per the dataset annotation $\vec{g}_a$, the centered base gaze direction $\vec{g}_b$ can be defined as:

\begin{equation}
  \vec{g}_b = \mathbf{^bT^{-1}_a} \  \vec{g}_a
\end{equation}

\subsection{Head Pose Distribution}
Given the application that the gaze estimation model will be used for, we need to define the expected pose of the user's head. In the case of our example case, we know the camera is mounted in front of and below the user, aiming 30 degrees up. We should thus expect the user's head to be centered on the image, but with a head pose pitch of 30 degrees up. In other words, the user is looking forward while the camera, positioned in front of the user but below them, looks up at the user's face with an angle of 30 degrees.

As this is the perfect case and not actually reflective of the reality, we will assume a variance of 10 degrees in both the head pitch and yaw. We thus define a bivariate normal distribution giving the head pitch $p$ and yaw $y$ as:

\begin{equation}
  \label{eq:bivariate_normal_aug}
  \begin{pmatrix}
      y \\
      p
      \end{pmatrix}\sim N\left(\begin{pmatrix}
      \mu_y = 0 \\
      \mu_p = 30
      \end{pmatrix},\begin{pmatrix}
      \sigma_y^2 = 10 & 0 \\
      0 & \sigma_p^2 = 10
  \end{pmatrix}\right)
\end{equation}

The \(\mu\) parameters are the means whereas the \(\sigma^2\) parameters are the variances. This distribution will be sampled when generating the augmented dataset to define the augmented head pose for each sample.

It should be noted in this case that the roll angle is not computed, as the normalization procedure that is generally used in gaze estimation models removes the roll component \cite{zhangRevisitingDataNormalization2018,suganoLearningbySynthesisAppearanceBased3D2014}. As such, there is no need to compute a roll just to remove it when normalizing the images.

\subsection{Sample Pose Transform}

By sampling the yaw and pitch from the previous distribution for each sample, we can calculate a rotation matrix to apply to the face mesh and gaze direction annotation. This computated rotation matrix will be the new augmented head pose, and thus we rotate the base face model and gaze direction using this rotation matrix.

We define the rotation $\mathbf{R}(y, p)$ around the pitch and yaw as:

\begin{multline}
  \mathbf{R}(y, p) = \mathbf{R_{pitch}}(p) \ \mathbf{R_{yaw}}(y) = \\
  \begin{bmatrix}
    1 & 0 & 0\\
    0 & \cos(p) & -\sin(p) \\
    0 & \sin(p) & \cos(p)
    \end{bmatrix}
    \begin{bmatrix}
      \cos(y) & 0 & \sin(y)\\
      0 & 1 & 0 \\
      -\sin(y) & 0 & \cos(y)
      \end{bmatrix}
\end{multline}

We then apply this rotation to the face mesh points $\mathbf{X_m}$ and the gaze direction $\vec{g}$ to generate our augmented sample from the dataset.

\begin{equation}
  \mathbf{X_m}' = \mathbf{R}(y, p) \ \mathbf{X_m}
\end{equation}

\begin{equation}
  \vec{g}' = \mathbf{R}(y, p) \ \vec{g}
\end{equation}

\begin{figure}
\centering{\includegraphics[width=\columnwidth]{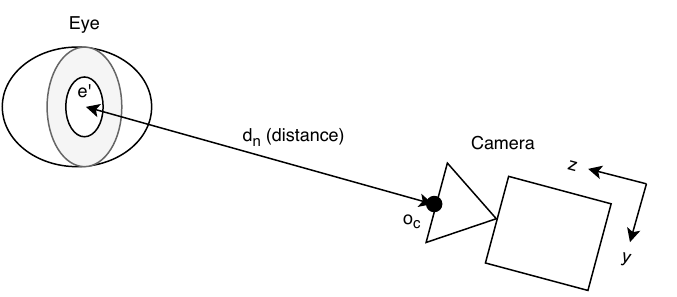}}
\caption{Geometric construction example of the virtual camera and the augmented eye position.}
\label{fig:geometric_augment_construction}
\end{figure}

\subsection{Mesh Projection and Normalization}
Using any rendering method, we can render the face mesh as an image, but to do so requires knowing the camera position and parameters, i.e., the field of view or focal length and the image pixel size. To simplify the workflow, we merge the normalization procedure often done as a preprocess step in gaze estimation \cite{suganoLearningbySynthesisAppearanceBased3D2014,zhangRevisitingDataNormalization2018} with the rendering of the face mesh.

Indeed, because we want our eye patch images to be normalized in distance, focal length and image pixel size, we use the normalized parameters to create a virtual camera viewpoint. 

For a given eye (left or right) and assuming a normalized distance $d_n$ and a normalized focal length $f_n$, we create a virtual camera situated at a $z$ distance of $d_n$ from the eye position on the transformed face mesh $e' = [e_x', e_y', e_z']$ with focal length $f_n$ (see Figure \ref{fig:geometric_augment_construction} for and example geometric construction). As such, the pose of the camera would point toward the position $z$ axis and its origin $o_c$ would be situated at:

\begin{equation}
  o_c = [e_x', e_y', e_z' - d_n]
\end{equation}

Since the rendering scene is now composed of the transformed face mesh $\mathbf{X_m'}$ and the virtual camera with focal length $f_n$, origin $o_c$ and aimed toward the positive $z$ axis, we can generate an eye patch image with the normalized image size required by our gaze estimation model. This can be done through any rendering tool, but we used Pyrender \cite{matlPyrender2022} to do so.

This rendering part should be done on the left eye and right eye if both are required as input for the gaze estimation model.

\subsection{Gaze Estimation Model Training}
Once every sample in the dataset has been augmented using the previous steps, we can use the samples as input to a gaze estimation model. In our case, we use the eye image patches (left and right) and the head pose $\mathbf{R}(y, p)$ as input to our model and we use the transform gaze direction $\vec{g}'$ to compute the loss. 

When evaluating the model or using it for inference, we simply use the normalization method described in \cite{zhangRevisitingDataNormalization2018} to generate the preprocessed input data and we denormalize the output gaze using the method described in the same paper. This has to be done because we do not generate a face mesh at evaluation time, but we use the actual image from the camera. Given that the training was done in a normalized space with the camera at a distance $d_n$ (in our case 600 mm) and a normalized focal length $f_n$ (in our case 650 px/mm), we must normalize the images to that space before using them as input to our trained model.

\section{Results and Discussion}
To analyse the performance of our dataset augmentation method, we trained a gaze estimation model based on a MobileNet architecture \cite{leblond-menardNonIntrusiveRealTime2023} on UTMultiview \cite{suganoLearningbySynthesisAppearanceBased3D2014}, MPIIFaceGaze \cite{zhangItWrittenAll2017} and GazeCapture \cite{krafkaEyeTrackingEveryone2016} with and without our augmentation method. Training for all datasets is done with the same hyperparameters that were tuned for the best accuracy \textit{before augmentation}. This should ensure that the improvement comes from the augmentation and not the choice of hyperparameters.

After training our model, we then evaluate it on a series of 69 samples from 3 different users taken on the setup as described by the example application (see Fig. \ref{fig:vis_scene}). The results are detailed in Table \ref{tab:accuracies}. The error values $e$ are given by computing the angle between the true gaze direction $\vec{g}$ and the estimated gaze direction $\vec{g}_e$:

\begin{equation}
    e = \ \cos^{- 1}\frac{| \vec{g} \cdot \vec{g}_{e} |}{|| \vec{g} || ||\vec{g}_{e} ||}
\end{equation}


\begin{table}[h]
  \renewcommand{\arraystretch}{1.3}
  \begin{threeparttable}
    \caption{Comparative results in gaze errors when trained on the original and augmented datasets}
    \label{tab:accuracies}
    \begin{tabular}{|l|l|l|l|}\hline
      Training Dataset & Original & Augmented & Reduction (\%)\\\hline
      UTMultiview\textsuperscript{*} & 9.2\degree & 6.9\degree & 25\% \\\hline
      MPIIFaceGaze & 9.5\degree & 3.0\degree & 68\% \\\hline
      GazeCapture & 6.0\degree & 3.0\degree & 49\% \\\hline
      \end{tabular}{}
    \begin{tablenotes}
      \small
      \item \textsuperscript{*}The UTMultiview dataset does not include full face images and thus it is not possible to use our own mesh generation technique and we therefore rely on the existing meshes contained in the dataset. 
    \end{tablenotes}
  \end{threeparttable}
\end{table}

\begin{figure}
\subfigure[Gaze pitch and yaw]{\includegraphics[width=.4\linewidth]{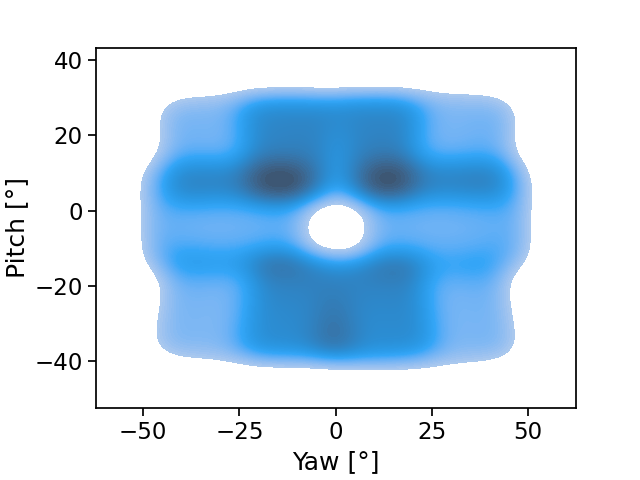}}
\hfill
\subfigure[Head pose pitch and yaw]{\includegraphics[width=.4\linewidth]{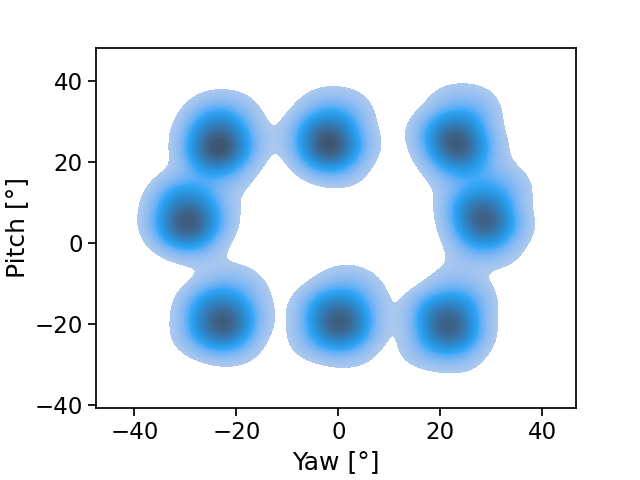}}
\caption{Distribution of the gaze direction and head pose pitch and yaw with respect to the camera's center axis for the UTMultiview dataset \cite{suganoLearningbySynthesisAppearanceBased3D2014}.} 
\floatfoot{Note that the gaze distribution in this case also includes sample where the user is looking above the camera, which is closer to our example application's average gaze direction.}
\label{fig:distribution_utmv}
\end{figure}

As can be seen from Table \ref{tab:accuracies}, the angular error decreases by 47\% on average. This is a significant increase in accuracy and leads to more consistent results in our sample application. This increase in accuracy does not significantly increase training time since it can be preprocessed once and trained on multiple times afterward without having to go through the augmentation steps again. It does not increase the development time significantly either since it has a very small amount of parameters to tune for, namely the bivariate normal distribution's means and variance, and the fact that they can be easily inferred from the application rather than a costly hyperparameter search.

The case of the UTMultiview \cite{suganoLearningbySynthesisAppearanceBased3D2014} dataset should also be noted, since the error reduction is lower for this dataset. Indeed, because the dataset does not include the full face in the images, it is thus not possible to generate the face mesh from our facial alignment method, which is also used in the evaluation step. Therefore, the base face pose used as the "zero" pose is not the same for both models. This leads to a discrepancy between the training data's annotations distribution and the evaluation's distribution.

Also, as can be seen from Fig \ref{fig:distribution_utmv}, the distribution of the gaze angles for the dataset also includes the user looking above the camera, which is not the case for MPIIFaceGaze \cite{zhangItWrittenAll2017} and GazeCapture \cite{krafkaEyeTrackingEveryone2016}. While the mean of the distribution might not coincide with the average gaze direction of the evaluation using our example application, it still includes more samples relevant to our application and thus it is to be expected that the UTMultiview-trained model will perform similarly with and without augmentation.

\section{Conclusion}

In conclusion, the developed augmentation method can be done quickly and leads to a significant (47\%) decrease in gaze estimation error in gaze estimation for assitive robotics tests, where three different users were asked to look at an object randomly placed in front of them over 69 samples. It does not require a costly hyperparameter search and can be used on existing datasets to adapt them to application-specific usages.

\bibliographystyle{IEEEtran} 
\bibliography{library}

\end{document}